# Signature Verification Approach using Fusion of Hybrid Texture Features

[a]Ankan Kumar Bhunia, [b]Alireza Alaei, [c]Partha Pratim Roy*
[a] Department of EE, Jadavpur University, India
[b] School of ICT, Griffith University, Australia
[c] Department of CSE, Indian Institute of Technology Roorkee, India
*email:proy.fcs@iitr.ac.in

*Abstract*

In this paper, a writer-dependent signature verification method is proposed. Two different types of texture features, namely Wavelet and Local Quantized Patterns (LQP) features, are employed to extract two kinds of transform and statistical based information from signature images. For each writer two separate one-class support vector machines (SVMs) corresponding to each set of LQP and Wavelet features are trained to obtain two different authenticity scores for a given signature. Finally, a score level classifier fusion method is used to integrate the scores obtained from the two one-class SVMs to achieve the verification score. In the proposed method only genuine signatures are used to train the one-class SVMs. The proposed signature verification method has been tested using four different publicly available datasets and the results demonstrate the generality of the proposed method. The proposed system outperforms other existing systems in the literature.

*Keywords—* Off-line signature verification; Texture features; Wavelet transform; Local phase quantization; Score level fusion.[1]

## I. Introduction

Human physiological or behavioural characteristic, biometrics, are commonly used for person identification / authentication in our daily life. Handwritten signature, as a unique human personal characteristic, is an accepted means of person authentication. However, manual handling of the large number of signatures generated in daily life is cumbersome. It demands an automatic algorithmic approach to deal with the problem of person verification based on handwritten signature [1-6]. As a result, many algorithms were developed in the literature to deal with the problem of signature-based person authentication in various applications including person identification and verification, crime detection, bank cheque fraud detection, etc. [2-6]. A signature verification method generally distinguishes between a person's original and forged signatures, accepting the original signatures and rejecting the forged ones. Three different types of forgeries namely random, simple and skilled were defined in the signature verification literature [1]. The skilled forgeries are generated by individuals who try to mimic the original signature and create one as close as possible to the original signature. Random and simple forgery samples are generated by individuals without any knowledge about the signers and their signatures. Indeed, the problem of signature verification considering skilled forgeries is a challenging task [1-2].

---

[1] Preprint Submitted

Signature-based person verification methods in the literature are categorized into: on-line and off-line approaches. On-line signature verification models use dynamic information, such as velocity, acceleration, pressure, stroke order, force, etc., whereas in off-line system signature images are the static source of information. Therefore, off-line signature verification is comparably more challenging with respect to on-line signature verification problem [1, 7].

In the past, Support Vector Machine has been found to be well suited for signature recognition [8-9]. Mainly, Binary class SVM (B-SVM) was used for signature modelling [9]. In B-SVM based signature verification methods, it is required to have both genuine and forged signatures for training. However, in practice only genuine signatures are available. In this work, a novel writer-dependent off-line signature verification model is proposed based on one-class Support Vector Machines (SVM). For each signature model two separate one-class SVMs are trained considering two different texture features extracted from genuine signatures. The scores obtained from the two one-class SVMs are then fused to verify the originality of the test signatures.

The main contributions of this work are as follows. First, a writer-dependent signature model by using one-class SVM is proposed that takes into account only genuine signatures for training, as it cannot be ensured that every writer in the system have examples from skilled forgeries. Thus, it makes our proposed model independent of skilled forgeries. Second, Wavelet and Local Quantized Patterns (LQP) features, as two different types of texture features, are employed to interpret two kinds of signature characteristics based on transform and statistical based features. Third, a novel score level classifier fusion method is employed to integrate the signature verification results obtained from two kinds of features, transform and statistical, and improve the performance of signature verification. Finally, a number of experiments using four different datasets collected in different contexts and languages is performed to evaluate the proposed signature verification system and to demonstrate the generality of the proposed technique.

The remainder of this paper is organized as follows. In Section II, the background of the work is reviewed. The proposed method is illustrated in Section III. Databases, evaluation metrics, experimental results, and comparative analysis are presented in Section IV. Finally, conclusions and future work are provided in Section V.

## II. A brief literature review

Signature verification methods in the literature follow a common pipeline composed of: i) pre-processing, ii) feature extraction, iii) training a classifier or creating a knowledge-based model, and iv) verification steps [1, 3]. The pre-processing includes various tasks, such as signature extraction, noise reduction, image normalization, binarization, skew/slant correction, and skeletonization [3]. All or a combination of these tasks are generally employed on signature images in each method to prepare to prepare them for the feature extraction step.

Following the pre-processing step, a set of discriminant features from the pre-processed signature images is extracted to interpret different aspects of the signatures for verification purposes. In the literature of offline signature verification, different feature extraction techniques including geometric, Connected Component (CC), directional and gradient, mathematical transformations, profiles and shadow-code, texture information, and interest points were proposed [7-8, 10-25]. Based on the level of granularity, feature extraction methods are grouped into local [9-13, 15-25, 33-35, 37-40, 52-53] and global [7-11, 16, 18, 31- 32, 35-36, 41] approaches.

Various methods were proposed in the literature of offline signature verification for creating signature models [2-6]. The created/trained models were then used to classify a test signature as a genuine or forged one. The signature models are either writer-dependent or writer-independent. In a writer-dependent approach, a specific signature model is created for each individual by using a few number of genuine signatures and random forgeries. In a writer-independent approach, however, a single model is created for all the individuals. Hybrid models were also proposed for signature verification [2-6]. Both writer-dependent and writer-independent models can be designed using machine learning, and similarity-based approaches. Neural Networks (NNs) [7, 12, 20, 39], Bayes classifier [10], Hidden Markov Models (HMMs) [5, 9, 24, 31, 33], Support Vector Machines (SVMs) [8, 9, 15-19, 21-22, 25, 36, 38-39, 41, 52], Gaussian Mixture Models (GMMs) [11], Gentle AdaBoost algorithm [37], and Ensembles of classifiers [38], as machine learning approaches, were used for signature verification in the past. As similarity-based approaches, different fuzzy membership functions (Takagi–Sugeno, trapezoidal, and triangular) [37, 42, 43-45, 49], K-Nearest Neighbour (KNN), Dynamic Time Warping, and point matching [10-11, 13, 17, 20, 23, 31-35] were also been developed in the literature for signature verification. Moreover, symbolic representation based approach was also employed for signature verification in the literature [19, 28, 53].

There are also many review papers in the literature to demonstrate the state-of-the-art methodological developments in the field of signature identification / verification [1-5]. A number of competitions were further organized to fairly evaluate the existing signature verification methods and report recent signature verification results and technological achievements [6, 14]. It is worth mentioning that literature of the offline signature verification is well established and a significant progress was demonstrated in this area. However, the problem of offline signature verification is still an open research problem [1-6, 14], as there are i) a low inter-class variability between every individual's genuine signatures and skilled forgeries, ii) a high intra-class variability in every individual's handwritten signatures compared to the other individual's biometrics, iii) a limited number of signatures for creating off-line signature verification models, and iv) only genuine signatures available for creating off-line signature verification models. Furthermore, we noted that only a few research works reported the use of hybrid features and score level fusion for signature verification in the literature. In this research work, we address the use of two different texture features and a score level fusion technique for the verification of off-line handwritten signatures. The proposed method also uses only genuine signatures for training purposes, as only genuine signatures are mostly available.

### III. Proposed Method

In our approach, two different types of texture features called Local Phase Quantization (LPQ) [48] and Discrete Wavelet Transform (DWT) [30] are used to characterise signature images. For each class of signatures two one-class SVMs are considered to compute two scores for a given signature using LPQ and DWT texture features. Then, a novel method of score level classifier fusion is applied to combine the two scores obtained from the one-class SVM classifiers and verify the authentication of the given input signature. The block diagram of the proposed system is given in the Fig. 1. The proposed system is divided into four major steps: a) pre-processing, b) feature extraction, c) one-class SVM classification, and d) fusion scores rule. Each step of the proposed method is detailed in the following subsections.

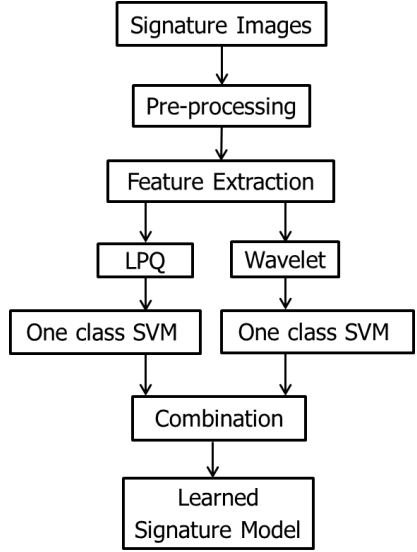

Fig. 1. Flowchart of our proposed framework

## 1. Pre-processing

Due to some unavoidable variations, in terms of size, pen-thickness, rotation and translation, in the signatures written by an individual, it is necessary to pre-process the signature images. The initial pre-processing task is the binarization of signature images. The Otsu's algorithm is applied and an optimal threshold is calculated separating the white pixels and black pixels so that their inter-class variance is maximal [46]. Then, a Gaussian filter is used to eliminate the noises from the input signature images and enhance the quality of the images. Finally, the signature images are cropped to make the features translation invariant. The cropped images are then used for feature extraction. Fig. 2 illustrates the different pre-processing tasks employed on an input signature image.

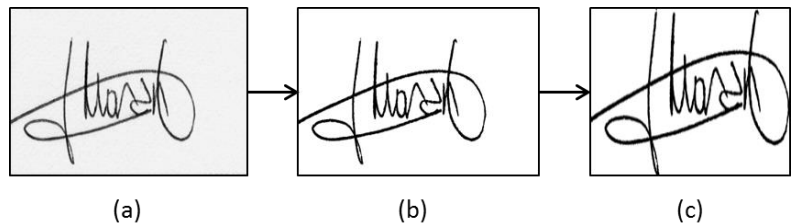

Fig. 2. (a) Original Image, (b) Image after binarization, (c) Image after filtering and cropping process

## 2. Feature extraction

In any pattern recognition problem, feature extraction is a crucial step. In our approach, two types of texture features: Local Phase Quantization (LPQ) [48] and Wavelet features (DWT) [30] are considered as feature extraction methods to extract two kinds of statistical and transform based features. LPQ is used for feature extraction in our proposed method, as it is a spatial blurring method, which is able to represent all spectrum characteristics of images in a very compact feature representation. The wavelet transform based features are further used to be complementary to the LPQ features.

## 2.1 Local Phase Quantization

LPQ is a blur intensive texture feature extraction method [48]. The spatial blurring of an image (g(x)) can be represented by a 2D-convolution between the original image (f(x)) and a point spread function or PSF (h(x)), where the vector x represent the coordinate $[x_1, x_2]^T$. The spatial blurring of f(x) can be expressed by a mathematical model [48] as follows.

$$g(x) = f(x) \otimes h(x) \quad (1)$$

In the frequency domain the convolution becomes a product operation [48] described as:

$$G(u) = F(u).H(u) \quad (2)$$

where u is a frequency and $G(u)$, $F(u)$, and $H(u)$ are the discrete Fourier transforms (DFT) of the blurred image (g(x)), the original image (f(x)) and the PSF (h(x)), respectively. Furthermore, if we consider the phase of the spectrum then the relation turns into a summation statement $\angle G = \angle F + \angle H$.

The magnitude and phase can be separated into two forms as demonstrated in the following.

$$|G(u)| = |F(u)|.|H(u)|) \text{ and } \angle G(u) = \angle F(u) + \angle H(u) \quad (3)$$

If h(x) is centrally symmetric, i.e. $h(x) = h(-x)$ then H is always a real value i.e. $\angle H \in \{0, \pi\}$.

For every pixel x from the image f(x), the local spectra are computed using a Short-Term Fourier Transform (STFT) in the local neighbourhood $N_x$ as follows:

$$F(u, x) = \sum_y f(y) \omega_R(y - x) e^{-j2\pi u^T y} \quad (4)$$

where $\omega_R(x)$ is a rectangular window function [48].

The local Fourier coefficients are computed at four low frequency components: $u_1 = [a, 0]^T$, $u_2 = [0, a]^T$, $u_3 = [a, 0]^T$, $u_4 = [a, -a]^T$, where a is small enough to satisfy $H(u_i) \geq 0$. For each point x we can write $F = [F(u_1, x), F(u_2, x), F(u_3, x), F(u_4, x)]$.

The phase information can be counted using a simple scalar quantization $q_j$.

$$q_j = \begin{cases} 1, & g_j \geq 0 \\ 0, & otherwise \end{cases} \quad (5)$$

where $g_j$ is the j-th component of the vector $(x) = Re\{F(x), Im\{F(x)\}\}$. Then, the label image is defined as:

$$f_{LPQ}(x) = \sum_{j=1}^{p} q_j(x) 2^{j-1} \quad (6)$$

Finally, from equation (6), a histogram of 256-dimensional feature vector is created and considered as LPQ features [48].

*2.2 Wavelet Features*

Discrete Wavelet divides a signal into different components in frequency domain [30]. The DWT of the signal x is calculated by employing a low pass and a high pass filter simultaneously on the input signal (Fig. 3).

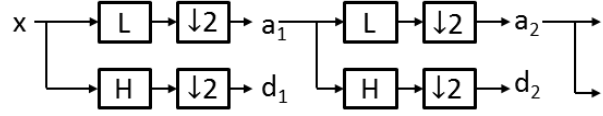

Fig. 3. Discrete Wavelet Transform (DWT) Tree

In the DWT tree presented in Fig. 3, H and L denote high pass and low pass filters, respectively. The symbol ↓2 denotes subsampling. Outputs of the filters are given by the following equations.

$$a_{j+1}[p] = \sum_{n=-\infty}^{\infty} l[n-2p]\, a_j[n] \tag{7}$$

$$d_{j+1}[p] = \sum_{n=-\infty}^{\infty} l[n-2p]\, d_j[n] \tag{8}$$

where $a_j$ is used for the next step of the transform and $d_j$ determines output of the transform. *l[n]* and *h[n]* are the coefficients of low and high pass filters, respectively.

In DWT, the input images are divided into four sub bands, i.e. LL, LH, HL, HH. LL is the average component or approximation image and LH, HL, HH are the three detail components. The LL sub band can be decomposed again and thereby producing more sub-bands [30]. This process can be carried out in many levels. The features obtained from these approximation and detail sub band images at different levels uniquely characterize a texture. The dimension of the extracted features is 120.

**3. One-class SVM**

One-class SVM is a special case of normal SVM that has been adapted to the one class classification problem [47]. It separates all the data points in a feature space from the origin and maximizes the distance from the hyper plane to the origin. The function returns +1 in a small region near the training data points and -1 otherwise. To separate the dataset from the origin, the following quadratic minimization function needs to be solved subjected to satisfying the other two conditions defined below.

$$\min_{w,\varepsilon_i,b} \frac{1}{2}\|w\|^2 + \frac{1}{\vartheta n}\sum_{i=1}^{n}\varepsilon_i - b \tag{9}$$

$$(w.\emptyset(x_i)) \geq b - \varepsilon_i \text{ and } \varepsilon_i \geq 0\ ;\ \text{For all i=1,…, n}$$

where $\vartheta$ is a tuning parameter, due to the importance of this parameter one-class SVM is often called as $\vartheta$-SVM. If $w$ and $b$ solve this problem, then the decision function, $f(x) = sign(w.\emptyset(x) - b)$ will be positive for most examples $x_i$ in the training set [47].

The function $K(x, x') = \emptyset(x)^T \emptyset(x')$ is known as kernel function. Popular choices for the kernel functions are linear, polynomial, sigmoidal but most used kernel function is Radial Base Function (RBF) shown in the following.

$$K(x, x') = \exp(-\frac{\|x-x'\|^2}{2\sigma^2}) \tag{10}$$

where the $\sigma \in R$ is a kernel parameter.

By using Lagrange technique and using a kernel function the decision function becomes:

$$f(x) = sign(\sum_{i=1}^{n} \alpha_i K(x, x') - b) \tag{11}$$

where $n$ is the number of training data and $\alpha_i$ is the Lagrange multiplier.

In our signature verification framework, the features extracted from the LPQ and DWT are fed into two separate one-class SVMs. Based on the training data, a specific model for the genuine signatures of each user is formed. The two trained one-class SVMs (for each individual) provide scores for a test signature and finally the two scores are combined to get a single verification score for the given test signature.

## 4. Combination rule for classification

Let T denote the target class for which we want to train our one-class SVM model and X represents an instance. The probability score of the instance X can be written as: $PS = P\left(\frac{T}{X}\right)$. A well-known and extensive approach for calculating the value $P\left(\frac{T}{X}\right)$ is to use a sigmoid function. Therefore, the probability score of a data point $x$ of a particular class j is given by:

$$PS_j(x) = sigmoid(\sum_{i=1}^{n} \alpha_{ij} K(x, x') - b_j) \tag{12}$$

$$sigmoid(x) = \frac{1}{1 + e^{-x}}$$

As in our proposed model two separate one-class SVMs are trained for each signature class using two types of features (LPQ and Wavelet), two probability scores $PS_j^{LPQ}(x)$ and $PS_j^{DWT}(x)$ are obtained for a signature $x$ of the signature class $j$. The average value of these two probability scores is considered to fuse the two classification probability scores and obtain a decision function $g_j(x)$, as it provides better verification results compared to other fusion strategies, such as maximum operator.

$$g_j(x) = \frac{PS_j^{LPQ}(x) + PS_j^{DWT}(x)}{2} \tag{13}$$

Finally, to classify genuine and forged signatures in the proposed model, the following decision rule is considered.

$$x = \begin{cases} Genuine, & if\ g_j(x) \geq T_j \\ Forgery, & otherwise \end{cases} \tag{14}$$

$T_j$ is an acceptance threshold which is defined as:

$$T_j = m_j + k\sigma_j \tag{15}$$

where $m_j$ and $\sigma_j$ are the respective mean and standard deviation computed from the decision function during the training phase. Moreover, $k$ is a control parameter, which is also needed to be tuned during the training phase to obtain optimal results.

## IV. Experimental results and discussion

### 1. Signature datasets and evaluation metrics

Four different offline signature databases, i.e. MCYT [29], GPDS-300 [49, 50], BHSig260 [51] and CEDAR [8], were used to evaluate the proposed signature verification method. The MCYT dataset is composed of 2250 signatures images from 75 signers and their associated skilled forgeries [29]. Each class contains 15 genuine signatures and 15 forgeries. Signatures are collected in the MCYT project by using an inking pen and paper.

The GPDS-300 dataset is composed of 16,200 offline signature images collected from 300 signers [49, 50]. Each signer has provided 24 genuine signatures. In each class of signatures, there are 30 skilled forged signatures obtained from 10 different forgers. Generally, the first 160 classes of the GPDS dataset were used for testing and the last 140 classes were used for tuning and training the parameters.

The BHSig260 dataset [51] contains 6,240 genuine signatures and 7,800 skilled forged signatures of which 100 sets of signatures were written in Bengali and the rest 160 sets were written in Hindi (two Indian languages). The handwritten offline signatures were collected from 260 individuals. Similar to the GPDS-300 database, each class contains 24 genuine and 30 skilled forged signatures. The collected data was scanned using a flatbed scanner with the resolution of 300DPI in grey scale and stored in TIFF format.

The CEDAR dataset [8] contains signatures of 55 signers. Each signer has provided 24 genuine signatures and 24 forged signatures. Hence the dataset is composed of 1,320 (55 × 24) genuine signatures as well as 1,320 forged signatures. To get an idea of signatures collected in each dataset, some samples of genuine and forged signatures from each dataset are shown in Table I.

For evaluating the performances of our system, we considered three commonly used error metrics in the literature called False Rejection Rate (FRR), False Acceptance Rate (FAR), and Average Error Rate (AER) [53].

Table I. Examples of genuine and forged signatures from four different datasets

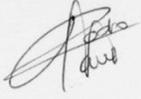

## 2. Experimental setting and results

The proposed signature verification method in this research work is dependent on three parameters, i.e. the percentage of outliers ($\vartheta$) and kernel parameter ($\sigma$) in the one-class SVM and also acceptance parameter *T*. To obtain optimal results, various couples of ($\vartheta, \sigma$) were considered during training the one-class SVMs using 8 genuine signatures per user for training and the best couple ($\vartheta_{opt} = 0.01, \sigma_{opt} = 0.01$) was selected where the Average Error Rate (AER) was the minimum. As the parameter *T* itself is dependent on the value *k*, the optimal values $\vartheta_{opt}$ and $\sigma_{opt}$ were used to tune the parameter *k*. The parameter *k* was tuned in such a way that the FRR and FER became equal to obtain an equal error rate (EER). The FRR and FAR values obtained using different values of *k* for GPDS-140 dataset are plotted in Fig. 4. As can be seen from Fig. 4, EER was obtained when the value of *k* was set to 2.18.

To demonstrate the sensitivity of the results obtained from other dataset to the parameter *k*, we further calculated the parameter *k* for all other datasets. The values of *k* for each dataset and their corresponding EER values obtained from the model considering 8 genuine signatures per user for training and the rest (16) of the genuine signatures and 30 forgeries for tuning the parameters are illustrated in the Table II. From Table II, we can note that the values of *k* for all the datasets are nearly the same. Thus, the proposed model is less dependent on the *k*-parameter on changing the dataset for training. As a results, in all the experiments on all the datasets the value of $k$ (=2.18) obtained from the GPDS-140 was considered for testing the proposed system.

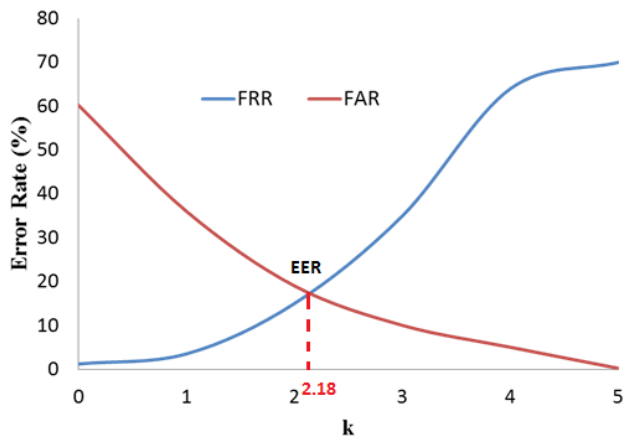

Fig. 4. FAR and FRR curves for different values of the parameter k on GPDS-140 dataset.

Table II. Results obtained from the proposed model using different k for different datasets

| Datasets | GPDS-300 | MCYT offline | BHsig-260 | CEDAR |
|---|---|---|---|---|
| k | 2.18 | 2.13 | 2.11 | 2.2 |
| EER (%) | 12.06 | 11.46 | 24.80 | 7.59 |

In our experiments we used four datasets -MCYT, GPDS-300, BHsig-260 and CEDAR. To evaluate our model several experiments were carried out using different number of samples from the four datasets considered for training and testing the proposed model. In order to compare our results to those reported in the literature, we trained our model with 4, 6, 8, 10, 12 genuine images per user. Table III represents different number of genuine signatures ($N_g$) and forgeries ($N_f$) used during the design and evaluation steps of the proposed signature verification model from each dataset.

Table III. Different number of samples used in design and evaluation steps

| Design Step | | | | | |
|---|---|---|---|---|---|
| Database | #Writers | Training (Samples per class) | | Tuning the parameters (Samples per class) | |
| | | $N_g$ | $N_f$ | $N_g$ | $N_f$ |
| GPDS-140 | 140 | 12, 10, 8, 6, 4 | 0 | 12, 14, 16, 18, 20 | 30 |
| MCYToffline | 40 | 12, 10, 8, 6, 4 | 0 | 3, 5, 7, 9, 11 | 15 |
| BHsig-260 | 100 | 12, 10, 8, 6, 4 | 0 | 12, 14, 16, 18, 20 | 30 |
| CEDAR | 20 | 12, 10, 8, 6, 4 | 0 | 12, 14, 16, 18, 20 | 24 |
| Evaluation Step | | | | | |
| Database | #Writers | Training (Samples per class) | | Testing (Samples per class) | |
| | | $N_g$ | $N_f$ | $N_g$ | $N_f$ |
| GPDS-160 | 160 | 12, 10, 8, 6, 4 | 0 | 12, 14, 16, 18, 20 | 30 |
| MCYToffline | 35 | 12, 10, 8, 6, 4 | 0 | 3, 5, 7, 9, 11 | 15 |
| BHsig-260 | 160 | 12, 10, 8, 6, 4 | 0 | 12, 14, 16, 18, 20 | 30 |
| CEDAR | 20 | 12, 10, 8, 6, 4 | 0 | 12, 14, 16, 18, 20 | 24 |

The error rates in terms of FRR, FAR, and AER obtained from the proposed signature verification model on the MCYT, GPDS, BHsig-260 and CEDAR datasets are shown in Table IV. As we have repeated ten times the training and the evaluation process with different randomly chosen samples, the average of FAR, FRR, AER are provided in Table IV. From Table IV we can see that the AERs gradually decrease when the number of signatures for training increases in all datasets. From Table IV we can further note that even when the number of training signatures is less our model provides good result on the CEDAR dataset.

Table IV. Results obtained from the proposed model considering different datasets for evaluation

| Dataset | Training | | Testing | | FAR (%) | FRR (%) | AER (%) |
| --- | --- | --- | --- | --- | --- | --- | --- |
| | $N_g$ | $N_f$ | $N_g$ | $N_f$ | | | |
| MCYT | 4 | 0 | 11 | 15 | 18.12 | 20.11 | 19.12 |
| | 6 | | 9 | | 13.55 | 16 | 14.78 |
| | 8 | | 7 | | 11.77 | 12.1 | 11.94 |
| | 10 | | 5 | | 8.78 | 10.23 | 9.5 |
| | 12 | | 3 | | 8.00 | 9.13 | 8.57 |
| GPDS-160 | 4 | 0 | 20 | 30 | 17.89 | 18.12 | 18.01 |
| | 6 | | 18 | | 15.89 | 15.18 | 15.54 |
| | 8 | | 16 | | 12.56 | 11.56 | 12.06 |
| | 10 | | 14 | | 10.89 | 9.53 | 10.26 |
| | 12 | | 12 | | 8.56 | 7.5 | 8.03 |
| BHsig-260 | 4 | 0 | 20 | 30 | 34.12 | 27.21 | 30.66 |
| | 6 | | 18 | | 27.12 | 26.12 | 26.62 |
| | 8 | | 16 | | 24.10 | 26.0 | 25.05 |
| | 10 | | 14 | | 20.1 | 24.18 | 22.14 |
| | 12 | | 12 | | 18.42 | 23.1 | 20.76 |
| CEDAR | 4 | 0 | 20 | 24 | 10.12 | 9.12 | 9.62 |
| | 6 | | 18 | | 8.2 | 8.4 | 8.3 |
| | 8 | | 16 | | 7.46 | 7.86 | 7.66 |
| | 10 | | 14 | | 6.12 | 7.2 | 6.66 |
| | 12 | | 12 | | 5.01 | 6.12 | 5.57 |

To demonstrate the effectiveness of our proposed fusion strategy compared to the signature verification using each feature set (LPQ or Wavelet) alone, we have also provided the results using the LPQ and wavelet features separately followed by one-class SVM for each dataset. The results are summarised in the Fig. 5, 6, 7, and 8. From the results plotted on Fig. 5, 6, 7, and 8, it can be noted that the LPQ features provide better results compared to the Wavelet features. However, the proposed score-level fusion method significantly increases (3% to 4%) the performance of the system compared to the systems using only the LPQ or Wavelet based features.

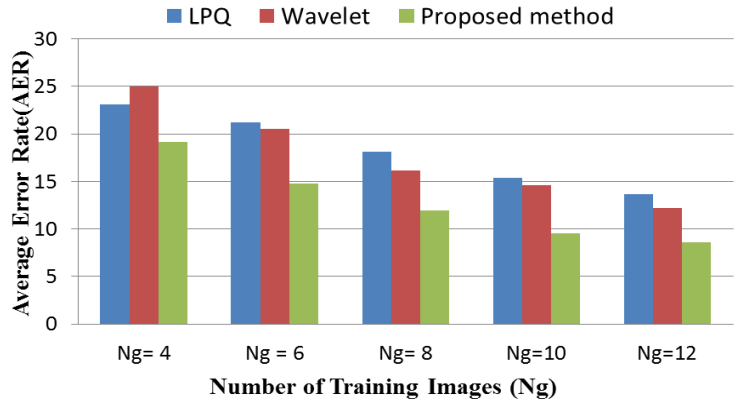

Fig. 5. The AERs obtained from the proposed model and using LPQ and Wavelet separately on MCYT Dataset.

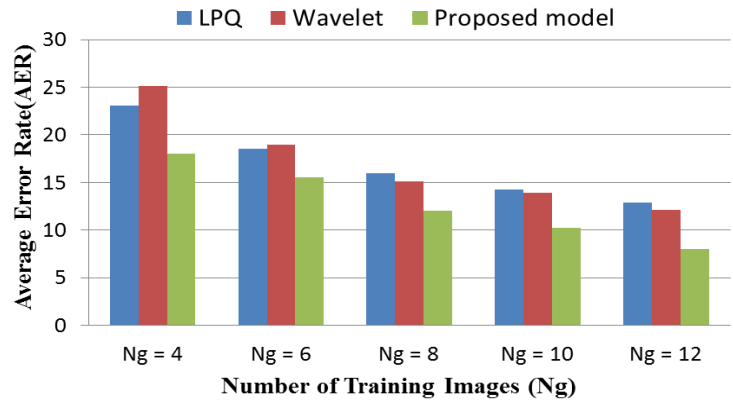

Fig. 6. The AERs obtained from the proposed model and using LPQ and Wavelet separately on GPDS Dataset.

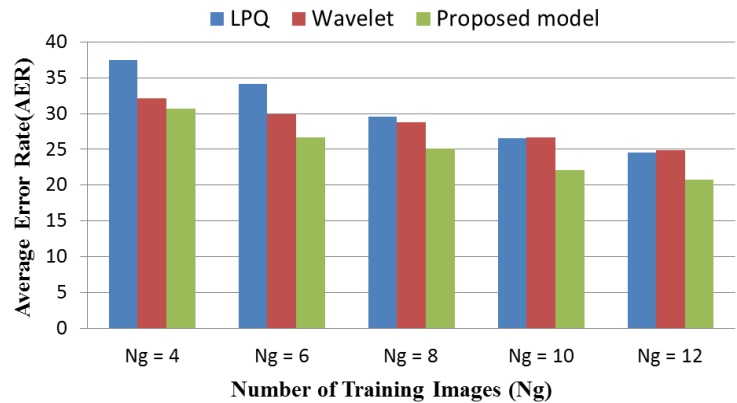

Fig. 7. The AERs obtained from the proposed model and using LPQ and Wavelet separately on BHsig-260 Dataset.

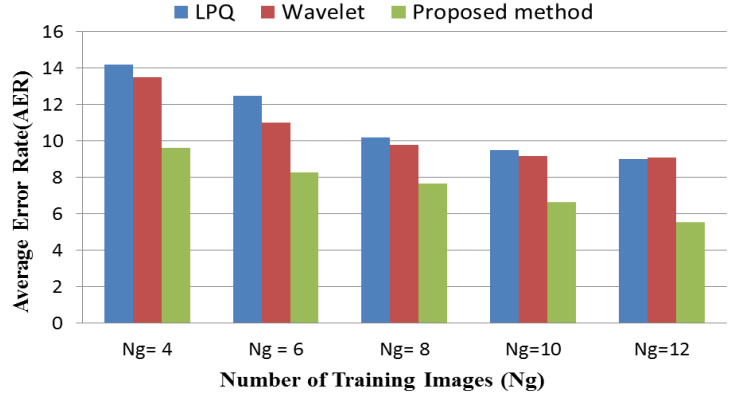

Fig. 8. The AERs obtained from the proposed model and using LPQ and Wavelet separately on CEDAR Dataset.

### 3. Comparative Analysis

A comparison of the results obtained from the proposed method and other approaches found in the literature on the MCYT, GPDS, CEDAR and BHsig260 datasets is provided in Table V. The results obtained based on the optimal $k$ in relation to each dataset in relation to the results computed based on $k = 2.18$ are also demonstrated in Table V. From the results shown in Table V, it can be noted that the proposed method outperforms the state-of-the-art methods in terms of AER when 8 or 12 signature samples were used for training the system. The results obtained when a lesser number of signature images (e.g. 4) were used for training are comparable with the state-of-the-art methods.

Table V. AER obtained from the proposed method compared to the other method. For each dataset, AER provided with the optimal $k$ of each dataset and general $k = 2.18$

| Dataset | Method | No. of Training Samples | AER (%) |
|---|---|---|---|
| MCYT | Vargas et al. [21] | 5<br>10 | 13.89<br>10.07 |
| | Alonso-Fernandez et al. [32] | 10 | 22.48 |
| | **Proposed method (k = 2.18)** | 4<br>10<br>12 | 19.11<br>9.50<br>8.56 |
| | **Proposed method (k = 2.13)** | 4<br>**10**<br>**12** | 19.16<br>**9.26**<br>**8.50** |
| GPDS-300 | Eskander et al. [37] | 12<br>14 | 15.24<br>13.96 |
| | Guerbai et. al. [36] | 4<br>8<br>12 | 16.92<br>15.95<br>15.07 |
| | Ferrer et al. [49] | **4**<br>8<br>12 | **16.10**<br>14.15<br>13.35 |
| | Alaei et al. [53] | 4<br>8<br>12 | 21.22<br>13.80<br>11.74 |

| | | 4 | 18.01 |
| --- | --- | --- | --- |
| | **Proposed method (k =2.18 )** | **8** | **12.06** |
| | | **12** | **8.03** |
| **BHsig260** | Alaei et al. [53] | 4 | 28.88 |
| | | 8 | 23.74 |
| | | 12 | 23.15 |
| | **Proposed method (k = 2.18)** | 4 | 30.66 |
| | | 8 | 25.05 |
| | | 12 | 20.76 |
| | **Proposed method (k = 2.11)** | 4 | 29.13 |
| | | 8 | 24.80 |
| | | **12** | **20.11** |
| **CEDAR** | Kumar et al. [26] | 16 | 6.02 |
| | Chen et al. [27] | 16 | 5.10 |
| | Guerbai et al. [36] | **4** | **8.70** |
| | | 6 | 7.83 |
| | | 12 | 5.60 |
| | **Proposed method (k = 2.18)** | 4 | 9.62 |
| | | 6 | 7.66 |
| | | 12 | 5.57 |
| | **Proposed method (k = 2.2)** | 4 | 9.50 |
| | | **8** | **7.59** |
| | | **12** | **5.50** |

## V. Conclusion

In this research work two types of texture features, LPQ and Wavelet, were employed to effectively characterise signature images. Two one-class SVMs were used to obtain two scores using the two feature sets extracted based on LPQ and Wavelet approaches. A score level fusion method was then proposed to combine two scores obtained from two types of information in order to achieve a higher signature verification performance. As one-class SVMs were used in classification step, only genuine signatures were used to train the proposed method. A wide range of experiments were conducted on four different datasets to evaluate the performance of the proposed method. The proposed method provided significantly better results compared to the state-of-the-art methods considering different off-line signature datasets.